\documentclass{article}
\usepackage{ijcai23}

\usepackage{times}
\usepackage{soul}
\usepackage{url}
\usepackage[hidelinks]{hyperref}
\usepackage[utf8]{inputenc}
\usepackage[small]{caption}
\usepackage{graphicx}
\usepackage{amsmath}
\usepackage{amsthm}
\usepackage{booktabs}
\usepackage{algorithm}
\usepackage{algorithmic}
\usepackage[switch]{lineno}
\usepackage{subfigure}
\usepackage{multirow}

\usepackage{amssymb}

\title{Multi-Scale Subgraph Contrastive Learning}

\author{
Yanbei Liu$^1$
\and
Yu Zhao$^2$\and
Xiao Wang$^{3}$\thanks{Corresponding author}\and
Lei Geng$^1$\And
Zhitao Xiao$^1$$^*$\
\affiliations
$^1$School of Life Sciences, Tiangong University\\
$^2$School of Electronics and Information Engineering, Tiangong University\\
$^3$School of Software, Beihang University\\
\emails
\{liuyanbei, zy0720, genglei, xiaozhitao\}@tiangong.edu.cn,
xiaowang@bupt.edu.cn
}

\begin{document}

\maketitle

\begin{abstract}
    Graph-level contrastive learning, aiming to learn the representations for each graph by contrasting two augmented graphs, has attracted considerable
    attention.
    Previous studies usually simply assume that a graph and its augmented graph as a positive pair, otherwise as a negative pair.
    However, it is well known that graph structure is always complex and multi-scale, which gives rise to a fundamental question: \textit{after graph augmentation, will the previous assumption still hold in reality?}
    By an experimental analysis, we discover the semantic information of an augmented graph structure may be not consistent as original graph structure, and whether two augmented graphs are positive or negative pairs is highly related with the multi-scale structures.  
    Based on this finding, we propose a multi-scale subgraph contrastive learning method which is able to characterize the fine-grained semantic information.
    Specifically, we generate global and local views at different scales based on subgraph sampling, and construct multiple contrastive relationships according to their semantic associations to provide richer self-supervised signals.
    Extensive experiments and parametric analysis on eight graph classification real-world datasets well demonstrate the effectiveness of the proposed method.
\end{abstract}

\section{Introduction}

Recently, graph neural networks (GNNs) have become a primary representation learning technique for dealing with many kinds of complex systems, ranging from the Internet and transportation graphs to biochemical interactions and social networks \cite{graphhulianwang,graphjiaotong,graphshejiao}.
Many real world applications usually require the graph-level representations, such as predicting molecular properties in drugs \cite{fenzi}, forecasting  protein functions in biological networks \cite{danbaizhi}, and predicting properties of circuits in circuit design \cite{dianlu}.
Therefore, GNNs, which are able to learn the graph-level representations, play an important role in these real applications.

Most existing GNNs belong to the supervised learning paradigm, which requires a lot of labeled graphs.
However, in many practical applications, collecting a large amount of labeled graph needs to consume a lot of resources.
For example, in the field of chemistry, properties of chemical molecules are often obtained through density functional theory calculations, which require expensive computational resources \cite{midufanhan}.
Therefore, Graph Contrastive Learning (GCL), one typical self-supervised paradigm, attracts considerable attention.
The general framework for GCL is to maximize the consistency of augmented views from the same anchor graph (positive pair), while minimizing the consistency of views from different anchor graphs (negative pair) \cite{graphcl,adgcl}.
Therefore, the key to graph contrastive learning is to ensure the semantic information matching between different augmented views, that is, views with similar semantics have similar representations.

There have been proposed many different graph augmentation strategies for GCL, e.g., node dropping \cite{graphcl}, edge perturbation \cite{adgcl}, attribute masking \cite{merit}, subgraph sampling \cite{graphcl}.
However, one fundamental question is that \textit{will the semantics of two augmented graphs still match in practice once the graph structure changes?}
It is well known that the graph structure is very complex, and different substructures may have their functional implications \cite{zijieguo,wangxiao1,wangxiao2}.
For example, in a social network, different communities may indicate factions, interest groups;  communities in a metabolic network might correspond to functional units, cycles, or circuits. Since the graph augmentation strategies essentially change the graph structures, it is hard to ensure the semantic information of different graph augmentations is matched.

Here, to provide more evidence for the above analysis, we perform an experiment to closely check the semantic relationship between different graph structures. Specifically, we select different substructures with different sizes on four real-world data, and then examine their semantic similarities (details can be seen in Section 2). The results clearly show that different graph structures have different semantics, and more importantly, the complex semantic information is positively related with the size of structure, i.e., larger subgraphs usually present larger semantic similarities. This well indicates that we cannot simply assume the semantic information of augmented structures are already matched, while more complex relationships between different augmented structures need to be carefully considered for an effective GCL, and forcibly requiring two augmented graphs with different semantics to be matched may largely mislead the GCL model.

In this paper, we propose a novel Multi-Scale SubGraph Contrastive Learning method, which models the multi-scale semantic information in different augmented subgraphs.
First, our experiment in Section 2 reveals that despite the graph structure is complex and different subgraphs have different semantics, their relationships can be generally divided into two facts: larger subgraphs, representing global view, usually have larger similarities, while smaller subgraphs, representing local view, usually have smaller similarities.
These findings motivate us to employ different learning strategies based on the two facts.
Specifically, we employ subgraph sampling to generate global and local views.
We expect to pull the global representations of same anchor graph close to each other, while also encouraging similarity between global and local views.
Meanwhile, we encourage the local representations to maintain a certain distance in the feature space.
Finally, we introduce a regressor to measure the similarity between local views to avoid the unreliability of traditional distance measures in high-dimensional spaces.

Our contribution can be summarized as follows:

\begin{itemize}
\item We study the roles of multi-scale augmented graphs in GCL and verify that the basic requirement on augmented subgraphs of GCL may not always hold in practice, i.e., not all the augmented subgraphs are semantically matched.

\item We propose a novel multi-scale subgraph contrastive learning method for graph-level representation learning.
Our model is able to consider multi-scale information of graph data and formulate different learning strategies according to its semantic associations.

\item We conduct comprehensive experiments on eight real-world datasets, and show that the proposed method achieves state-of-the-art performance on both unsupervised and semi-supervised graph classification tasks.

\end{itemize}

\section{An Experimental Investigation}
In this section, we employ data augmentation to obtain information at different scales in the graph dataset, and analyze its semantic similarity, aiming to obtain the semantic association between information at different scales.
We take four molecule graph dataset (MUTAG, NCI1, DD, PROTEINS) in the TUDataset \cite{tudataset} as examples.
First, we trained a 5-layer graph isomorphic network \cite{gin} with a hidden dimension of 32 on a single dataset via the Adam \cite{adam} optimizer in a supervised training paradigm.
Second, we augment the original graph with a subgraph sampling strategy based on random walks, and generate subgraphs at different scales by controlling the number of nodes in the augmented view \cite{graphcl}.
Since the starting point of the random walk is randomly selected, the generated subgraph is not fixed.
We generate two augmented views for each graph in the dataset, and treat the augmented views from the same original graph as subgraph pair.
Then, we remove the classifier of the graph isomorphic network, input subgraph pair generated by each graph into the network to obtain its feature vectors, and calculate the cosine similarity between the two feature vectors. 
We do the same for every graph in the dataset.
Finally, we take this similarity as the semantic similarity between views and compute the mean and variance of the semantic similarity between subgraph pairs generated from all original graphs in this dataset.
\begin{figure}[t]
\centering
\subfigure[MUTAG]{
\begin{minipage}[t]{0.48\linewidth}
\centering
\includegraphics[scale=0.25]{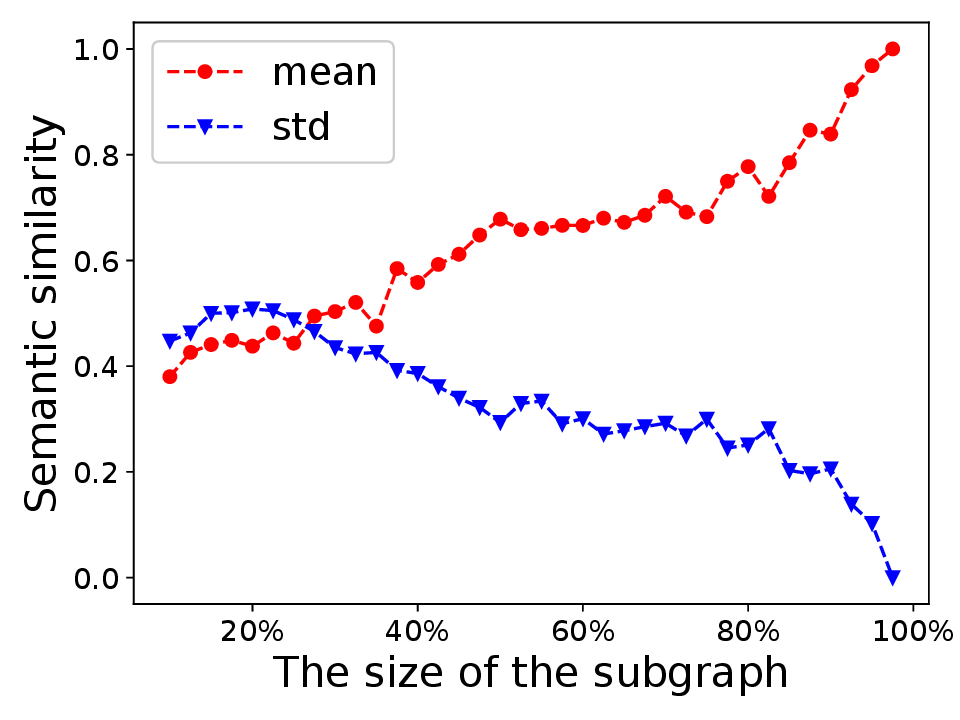}
\end{minipage}%
}%
\subfigure[NCI1]{
\begin{minipage}[t]{0.48\linewidth}
\centering
\includegraphics[scale=0.25]{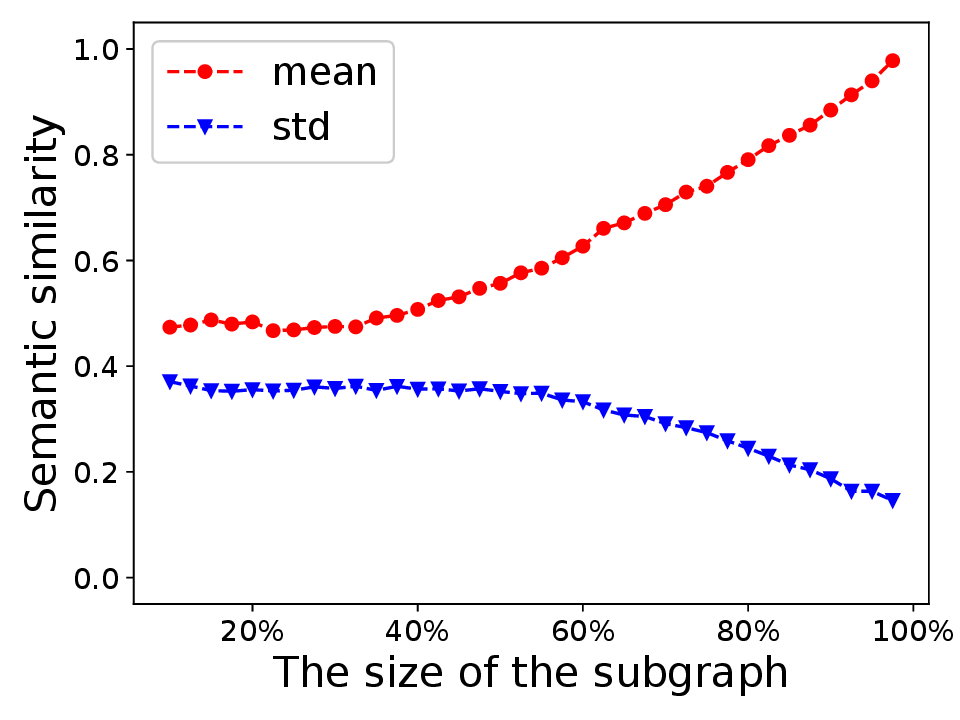}
\end{minipage}%
}%

\centering
\subfigure[DD]{
\begin{minipage}[t]{0.48\linewidth}
\centering
\includegraphics[scale=0.25]{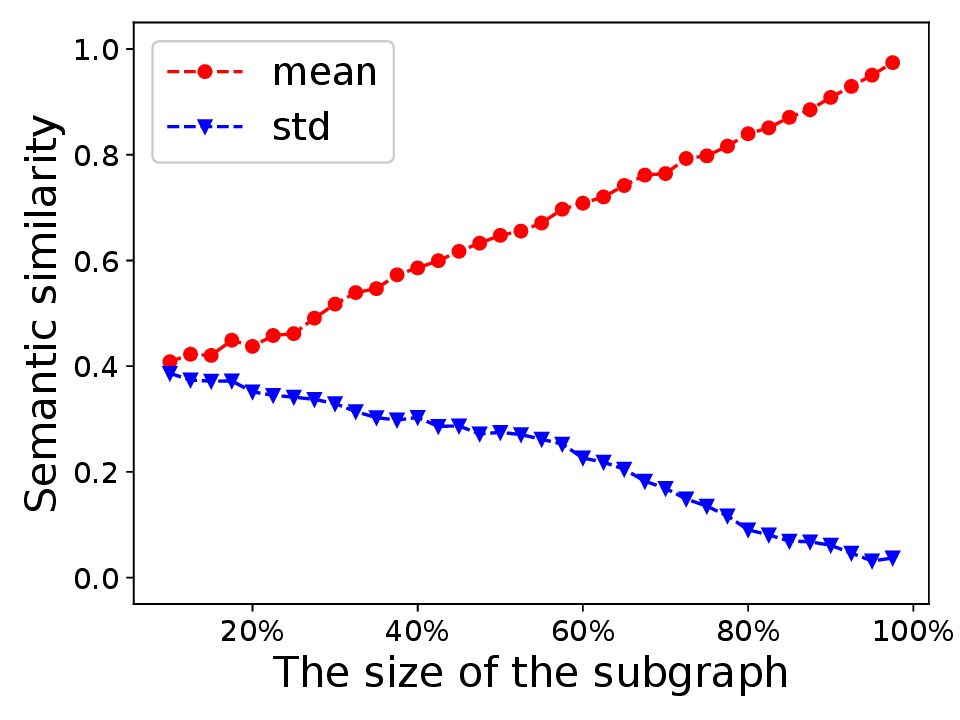}
\end{minipage}%
}%
\centering
\subfigure[PROTEINS]{
\begin{minipage}[t]{0.48\linewidth}
\centering
\includegraphics[scale=0.25]{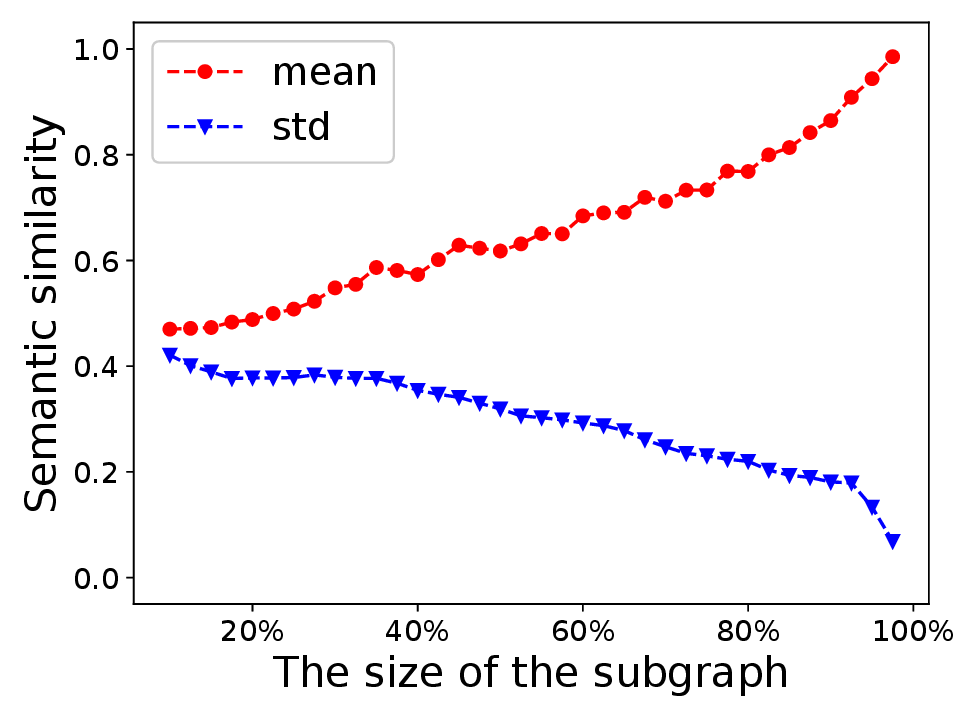}
\end{minipage}%
}%
\caption{Semantic similarity of subgraphs of different sizes.}
  \label{fig1}
\end{figure}

Figure \ref{fig1} shows the semantic similarity between subgraph pairs at different scales.
It can be seen that, firstly, as the size of subgraph increases, the mean value of the semantic similarity between the generated subgraph pairs increases continuously, which means the content that the larger subgraph describe is more similar on a semantic level.
Secondly, the variance of the semantic similarity between subgraph pairs is decreasing, which means that the content changes of their descriptions are also decreasing.
The above phenomenon implies that small-scale subgraphs always describe different content in the graph and vary greatly, while large-scale subgraphs describe more similar content, so the semantic association between view pairs at different scales is not uniform.
Current GCL methods usually perform simple perturbations on the graph structure to obtain augmented views and consider them to have similar semantics. 
However, the semantic information of augmented views at different scales is different, which requires us to distinguish them at a finer granularity.

\begin{figure*}[t]
    \centering
    \includegraphics[scale=0.45]{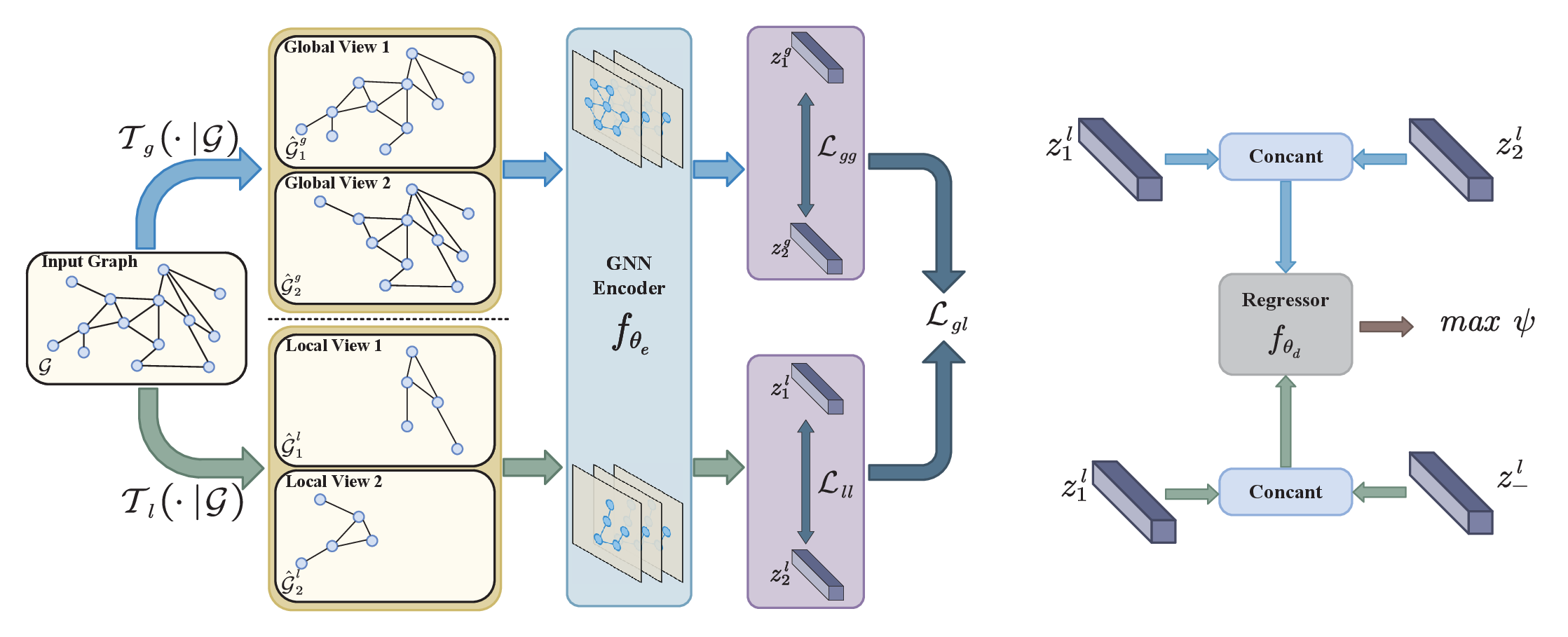}
    \caption{The overall architecture of MSSGCL (left).
The original graph generates global views and local views pairs through random walks with controlled number of nodes, which are then fed into the encoder to obtain global and local representations.
We maximize the similarity between global views and the similarity between global and local views by optimizing $\mathcal{L}_{gg}$ and $\mathcal{L}_{gl}$.
The dissimilarity between local views is encouraged by optimizing the output of a learned similarity measure $\mathcal{L}_{ll}$.
$f_{\theta _e}$ is the GNN-based encoder, which includes a backbone network followed by a multi-layer perceptron.
$f_{\theta _d}$ (right) is a learnable regressor to measure the similarity between local views.
}
    \label{fig2}
\end{figure*}

\section{Methodology}
\paragraph{Problem definition.}
Given an undirected graph $\mathcal{G}=\left\{ \mathcal{V},\mathcal{E} \right\} $ where $\mathcal{V}$ denotes the set of $\left| \mathcal{V} \right|$ nodes and $\mathcal{E}=e_{ij}\in \mathbb{R}^{\left| \mathcal{V} \right|\times \left| \mathcal{V} \right|}$ indicates the adjacency matrix where each entry $e_{ij}$ is the linkage relation between nodes $i$ and $j$.
For self-supervised graph representation learning, given a bunch of unlabeled graphs $G=\left\{ \mathcal{G}_1,\mathcal{G}_2,...,\mathcal{G}_M \right\} $, our goal is to learn a representation vector $z_i$ for each graph  through supervisory signals obtained from the data itself.
The resulting representation vector $z_i$ can be used in different types of downstream tasks, such as graph classification.

\paragraph{Overall framework.}
In this section, we will introduce our multi-scale subgraph based graph contrastive learning framework.
As shown in Figure \ref{fig2}, it consists of three major components, including generation of multi-scale subgraphs, graph-level representation learning and multi-scale contrastive loss.
Given a $\mathcal{G}$, we first apply graph augmentation techniques to obtain two sets of subgraphs with different sizes, namely global views  and local views.
Then, we utilize GNN-based encoders to learn their representations $z^g$ and $z^l$, respectively.
Subsequently, we perform the hierarchical local-global contrastive learning to optimize model parameters by the gradient descent.

\subsection{Generation of Multi-Scale Subgraphs}
In this section, we focus on graph-level data augmentation to obtain multi-scale subgraphs.
Given a graph $\mathcal{G}\in \left\{ \mathcal{G}_m:m\in M \right\} $, we define the augmented graph as $\hat{\mathcal{G}}\sim \mathcal{T}\left( \hat{\mathcal{G}}\left| \mathcal{G} \right. \right) $, where $\mathcal{T}\left( \cdot \left| \mathcal{G} \right. \right) $ is a predefined augmentation over the original graph, representing the human prior knowledge of graph data.
Generally speaking, there are four common graph augmentation methods, namely node dropping, attribute masking, edge perturbation and subgraph sampling.
Previous study \cite{graphcl} has shown that, compared with attribute masking and edge perturbation, subgraph sampling can benefit downstream tasks across different categories of graph datasets.
So we mainly employ subgraph sampling as data augmentation strategy.
As mentioned above, there are large differences in semantic similarity between pairs of subgraphs with different scales.
Large-scale subgraph pairs have high semantic similarity and small variance, while small-scale subgraph pairs have the opposite.
Therefore, we can form local and global subgraph sampling strategies $\mathcal{T}_l\left( \cdot \left| \mathcal{G} \right. \right) $ and $\mathcal{T}_g\left( \cdot \left| \mathcal{G} \right. \right) $ by limiting the scale of generated views. 
Then we perform the two subgraph sampling strategies on the given graph $\mathcal{G}$, and obtain two global views $\left\{ \hat{\mathcal{G}}_{i}^{g} \right\} _{i=1}^{2}$ and two local views $\left\{ \hat{\mathcal{G}}_{i}^{l} \right\} _{i=1}^{2}$
 for a single graph.

\subsection{Graph-level Representation Learning}
After acquiring the different scale augmentation views, we further learn their latent representations.
Here we employ GNNs as the encoder to obtain node representations by iteratively aggregating neighbor information.
Next, we will take the global view $\hat{\mathcal{G}}_{i}^{g}$ as an example to illustrate the representation learning process, which is exactly the same for the local view.
Given an augmented graph $\hat{\mathcal{G}}_{i}^{g}$ with its feature matrix $X\in \mathbb{R}^{\left| \mathcal{V} \right|\times N}$, where $x_n=X\left[ n,: \right] ^{\text{T}}$ is the $N$-dimensional feature vector of node $v_n$.
In general, in a $K$-layer GNN, the node representations of $k$-layer can be formalized as:
\begin{equation}
\begin{aligned}
    a_{n}^{\left( k \right)}=AGGREGATE^{\left( k \right)}\left( \left\{ h_{u}^{\left( k-1 \right)}:u\in \mathcal{N}\left( n \right) \right\} \right),\\
    h_{n}^{\left( k \right)}=COMBINE^{\left( k \right)}\left( h_{n}^{\left( k-1 \right)},a_{n}^{\left( k \right)} \right),
\end{aligned}
\end{equation}
where $h_{n}^{\left( k \right)}$ is the representation of node $v_n$ at $k$-th layer, with $h_{n}^{\left( 0 \right)}=x_n$. $\mathcal{N}\left( n \right) $ is the set of neighbors of node $v_n$.
The $AGGREGATION^{\left( k \right)}\left( \cdot \right)$ and $COMBINE^{\left( k \right)}\left( \cdot \right)$ are important functional components of GNN.
After the $K$-layer propagation, the $READOUT$ function aggregates the feature vectors of all nodes to get the representations of the entire graph which can be used for downstream tasks:
\begin{equation}
f\left( \hat{\mathcal{G}}_{i}^{g} \right) =READOUT\left( \left\{ h_{n}^{\left( K-1 \right)}:v_n\in \mathcal{V}_i \right\} \right) .
\end{equation}

Similar to contrastive learning in the computer visual domain \cite{simclr}, a non-linear transformation $g\left( \cdot \right) $ is used to map graph-level representations into the latent space to enhance the performance:
\begin{equation}
z_{i}^{g}=g\left( f\left( \hat{\mathcal{G}}_{i}^{g} \right) \right) .
\end{equation}

Through a similar procedure, we can obtain the representations of the local view as follows:
\begin{equation}
z_{i}^{l}=g\left( f\left( \hat{\mathcal{G}}_{i}^{l} \right) \right) .
\end{equation}

\subsection{Multi-Scale Contrastive Loss}
There are significant differences in the semantic similarity between subgraph pairs at different scales.
Existing methods may suffer from some drawbacks in applying contrastive learning between subgraph pairs.
For example, GraphCL directly pulls the representations distance between small-scale subgraphs.
Direct application of such existing contrastive learning strategies creates noisy and potentially contradictory constraints that complicate the learning process and affect performance.
To address this, we introduce two subgraph pairs of different sizes by limiting the number of nodes, namely global view and local view.
After that, we optimize the global-to-global, local-to-global and local-to-local relationships, respectively.
In the following, we denote $l_s$ as a general contrastive loss, which is introducted as the noise-contrastive estimation loss \cite{infonce}, and we take the global representation as an example to illustrate its optimization process:
\begin{equation}
l_s\left( z_{1}^{g},z_{2}^{g} \right) =-\log \frac{\exp \left( z_{1}^{g}\cdot z_{2}^{g}/\tau \right)}{\exp \left( z_{1}^{g}\cdot z_{2}^{g}/\tau \right) +\sum{\exp \left( z_{1}^{g}\cdot z_{-}^{g}/\tau \right)}},
\end{equation}
where $z_{1}^{g}$ and $z_{2}^{g}$ are global representations from the same graph, $z_{-}^{g}$ denotes the negative samples, which can be seen as global representations from other graphs in our architecture and $\tau $ is a temperature hyperparameter.

\paragraph{Global-to-global.}
Since the global view pair contains most of the content of the original graph, it owns similar  semantic information.
Our goal is to maximize the similarity of the global representations from the same original graph and minimize the similarity of the global representations from different original graphs.
The global-to-global loss can be written as follows:
\begin{equation}
\mathcal{L}_{gg}=\mathbb{E}_{p\left( z^g \right)}\left[ l_s\left( z_{1}^{g},z_{2}^{g} \right) \right],
\end{equation}
where $p\left( z^g \right) $ is the distribution of $z^g$.

\paragraph{Global-to-local.}
The global view owns a subgraph with large size, so it contains the content of the local view to a large extent, which ensures that the global view can share some semantic information with the local view.
\begin{algorithm}[t]
    \caption{The training process of the MSSGCL}
    \label{alg1}
    \textbf{Input}: Training set $D=\left\{ \mathcal{G}_i \right\} _{i=1}^{M}$, batch size $P$, training epochs $T$, a GNN based encoder $f$, global and local augmentation distributions $\mathcal{T}_g$ and $\mathcal{T}_l$\\
    \textbf{Output}:The pre-trained GNN encoder $f_{\theta _e}$
    
    \begin{algorithmic} 
        \STATE Initialize GNN encoder and regressor parameters 
        \WHILE{$t < T$}
        \STATE Sample graph minibatch $B_{\mathcal{G}}$ from $D$
        \STATE $B_{\hat{\mathcal{G}}^g}\sim \mathcal{T}_g\left( B_{\hat{\mathcal{G}}^g}\left| B_{\mathcal{G}} \right. \right) $, $B_{\hat{\mathcal{G}}^l}\sim \mathcal{T}_l\left( B_{\hat{\mathcal{G}}^l}\left| B_{\mathcal{G}} \right. \right) $
        \STATE $z^g,z^l\gets Eqs.\left( 1,2,3,4 \right) $
        \FOR{\textbf{all} $i\in \left\{ 1,...,P \right\} $ }
        \STATE Get positive local view pair $z_{1}^{l},z_{2}^{l}$
        \STATE Randomly choose a local view $z_{-}^{l}$
        \ENDFOR
        \STATE Update $\theta _d$ to maximize the $\psi \left( z_{1}^{l},z_{2}^{l}  \right)$ via Eq.(\ref{cost})
        \FOR{\textbf{all} $i\in \left\{ 1,...,P \right\} $ }
        \STATE Calculate similarity loss of global views
        \STATE Calculate similarity loss of global and local views
        \STATE Calculate similarity loss of local views
        \ENDFOR
        \STATE Update $f_{\theta _e}$ to minimize the total loss by Eq.(\ref{eq9})
        \ENDWHILE
    \end{algorithmic}
\end{algorithm}
\begin{table}[h]
    \begin{center}
    \resizebox{\linewidth}{!}{
            \begin{tabular}{c|c|c|c|c}
            \toprule
                Dataset & Category & Graph& Node& Edge \\
                \midrule
                 MUTAG & Molecules & 188 & 17.93 & 19.79 \\
                 NCI1 & Molecules & 4110 & 29.87 & 32.30 \\
                 PROTEINS & Molecules & 1113 & 39.06 & 72.82 \\
                 DD & Molecules & 1178 & 284.32 & 715.66 \\
                \midrule
                 IMDB-B & Social Network  & 1000 & 19.77 & 95.63 \\
                 COLLAB & Social Network  & 5000 & 74.49 & 2457.78 \\
                 RDT-B & Social Network  & 2000 & 429.63 & 497.75 \\
                 RDT-M5K & Social Network  & 5000 & 508.52 & 594.87 \\
                \bottomrule
            \end{tabular}
            }
        \caption{Statistics of datasets.}
        \label{tab1}
    \end{center}
\end{table}
Therefore, we define a loss function that pulls to narrow the distance between local and global representations in the latent space and establish the connection between the local and global  representations, which can be written as follows:
\begin{equation}
\mathcal{L}_{gl}=\mathbb{E}_{p\left( z^g,z^l \right)}\left[ \sum_{i=1}^2{\left( l_s\left( z_{i}^{g},z_{1}^{l} \right) +l_s\left( z_{i}^{g},z_{2}^{l} \right) \right)} \right] ,
\end{equation}
where $p\left( z^g,z^l \right) $ is the joint distribution of $z^g$ and $z^l$.
\begin{table*}[t]
    \centering
    \resizebox{\textwidth}{!}{
    \begin{tabular}{c|ccccccccc}
        \toprule
        Method & NCI1 & PROTEINS & DD & MUTAG & COLLAB& RDT-B & RDT-M5K & IMDB-B & AVG. \\
        \midrule
                 WL & 80.01$\pm$0.50  & 72.92$\pm$0.56 & 74.02$\pm$2.28 & 80.72$\pm$3.00  & 60.30$\pm$3.44 & 68.82$\pm$0.41  & 46.06$\pm$0.21  & 72.30$\pm$3.44 & 70.52  \\ 
                 DGK & 80.31$\pm$0.46 & 73.30$\pm$0.82 & 74.85$\pm$0.74 & 87.44$\pm$2.72 & 64.66$\pm$0.50 & 78.04$\pm$0.39 & 41.27$\pm$0.18 & 66.96$\pm$0.56  & 70.85\\ 
                 \midrule
                 sub2vec & 52.84$\pm$1.47 & 53.03$\pm$5.55 & 54.33 $\pm$2.44 & 61.05$\pm$15.80 & 55.26$\pm$1.54& 71.48$\pm$0.41 & 36.68$\pm$0.42 & 55.26$\pm$1.54 & 55.04\\
                 node2vec & 54.89$\pm$1.61 & 57.49$\pm$3.57 & 74.77$\pm$0.51 & 72.63$\pm$10.20 & 54.57$\pm$0.37& 72.76$\pm$0.92 & 31.09$\pm$0.14 & 38.60$\pm$2.30 & 57.10\\
                 graph2vec & 73.22$\pm$1.81  & 73.30$\pm$2.05 & 70.32$\pm$2.32 & 83.15$\pm$9.25 & 71.10$\pm$0.54 & 75.48$\pm$1.03 & 47.86$\pm$0.26 & 71.10$\pm$0.54 & 70.69\\
                 \midrule
                 InfoGraph & 76.20$\pm$1.06  & 74.44$\pm$0.31 & 72.85$\pm$1.78 & 89.01$\pm$1.13 & 70.65$\pm$1.13 & 82.50$\pm$1.42 & 53.46$\pm$1.03 & 73.03$\pm$0.87 &74.02\\
                 GraphCL & 77.87$\pm$0.41  & 74.39$\pm$0.45 & 78.62$\pm$0.40 & 86.80$\pm$1.34 & 71.36$\pm$1.15 & 89.53$\pm$0.84 & 55.99$\pm$0.28 & 71.14$\pm$0.44 & 75.41\\
                 JOAO   & 78.07$\pm$0.47 & 74.55$\pm$0.41 & 77.32$\pm$0.54 & 87.35$\pm$1.02 & 69.50$\pm$0.36 &85.29$\pm$1.35 & 55.74$\pm$0.63 & 70.21$\pm$3.08  &74.75\\
                 JOAO V2 & 78.36$\pm$0.53 & 74.07$\pm$1.10 & 77.40$\pm$1.15 & 87.67$\pm$0.79 & 69.33$\pm$0.34 &86.42$\pm$1.45 & 56.03$\pm$0.27 & 70.83$\pm$0.25 &75.01\\
                 SimGRACE & 79.12$\pm$0.44 & 75.35$\pm$0.09 & 77.44$\pm$1.11 & 89.01$\pm$1.31 & 71.72$\pm$0.82 &89.51$\pm$0.89 & 55.91$\pm$0.34 & 71.30$\pm$0.77 & 76.17\\
                 \midrule
                 MSSGCL & \textbf{81.45 $\pm$ 0.48} & \textbf{75.49 $\pm$ 0.70} & \textbf{79.73 $\pm$ 0.44} & \textbf{89.68 $\pm$ 0.57} &\textbf{73.48 $\pm$ 0.83} & \textbf{91.08$\pm$0.78} & \textbf{56.17$\pm$0.18} & \textbf{73.14$\pm$0.38}  &\textbf{77.52}\\
        \bottomrule
    \end{tabular}
    }
    \caption{Comparison of classification accuracy with other baselines in unsupervised setting. AVG. denotes the average accuracy.}
        \label{tab2}
\end{table*}
\paragraph{Local-to-local.}
Two local views from the same original graph usually describe different contents with low semantic similarity.
Thus, instead of similarity of local representations as most existing studies have done, we encourage their dissimilarity, making them farther apart in the representation space.
Given a measure function $l_d$, we express maximizing the dissimilarity between local views as minimizing the loss:
\begin{equation}
\mathcal{L}_{ll}=\mathbb{E}_{p\left( z^l \right)}\left[ l_d\left( z_{1}^{l},z_{2}^{l} \right) \right].
\end{equation}
In principle, we can choose any similarity measurement method, such as cosine similarity, but the high dimension of feature space may lead to the learning of meaningless representations \cite{gaoweikongjian}, and the semantic relationship between different local view pairs varies greatly.
Therefore, instead of using a traditional metric to push local views away from each other, we measure the similarity of local view pairs through a regressor with learnable parameters.

Specifically, we exploit the intuition that although local views from different graphs may contain the same semantic content, in general we still expect local views from the same graph to be more closely related to each other than that from different graphs.
To implement this expectation, we utilize a learnable regressor $f_{\theta _d}:=\mathbb{R}^n\times \mathbb{R}^n\rightarrow \mathbb{R}^+$ that gives a similarity measure between local views.
The parameters of the regressor $\theta _d$ can be trained in conjunction with the parameters of the encoder.
Therefore, we train the regressor by maximizing the cost function:
\begin{equation}
\psi \left( z_{1}^{l},z_{2}^{l} \right) =\mathbb{E}_{p\left( z_{1}^{l},z_{2}^{l} \right)}\left[ f_{\theta _d}\left( z_{1}^{l},z_{2}^{l} \right) \right] -\mathbb{E}_{p\left( z_{1}^{l}\otimes z_{-}^{l} \right)}\left[ f_{\theta _d}\left( z_{1}^{l},z_{-}^{l} \right) \right] ,
\label{cost}
\end{equation}
where $z_{-}^{l}$ is the negative sample of $z_{1}^{l}$, i.e., the local representations from different graphs, which can be obtained by random sampling from the same batch.
And $p\left( z_{1}^{l}\otimes z_{-}^{l} \right) $ is the product of two marginal distributions.
After that, we take the trained regressor as our metric function:
\begin{equation}
\ell _d=f_{\theta _{d}^{'}},\,\,s.t.\,\,\theta _{d}^{'}=\underset{\theta _d}{arg\max}\psi \left( z_{1}^{l},z_{2}^{l} \right).
\end{equation}

In general, we use the regressor to make local representations from the same graph more similar than those from different graphs.
Then, we train the encoder to minimize the metric value between local representation pairs from the same graph to account for their low semantic similarity.

In summary, our model can be summarized as a bi-level iterative optimization problem and the final loss function can be written as follows:
\begin{gather}
    \min \mathcal{L}_{gg}+\lambda _1\mathcal{L}_{gl}+\lambda _2\mathcal{L}_{ll},
    \label{eq9}
\end{gather}
where $\lambda _1$ and $\lambda _2$ are hyper-parameters to balance different loss terms.
The implementation details of our framework are provided in Algorithm \ref{alg1}.
\section{Experiment}
In this section, we compare the proposed method with other advanced models in unsupervised and semi-supervised learning settings to evaluate its performance.
Furthermore, we perform ablation experiments to demonstrate the effectiveness of various components of the proposed method.
\subsection{Setup}
\paragraph{Datasets.}
We adopt the TUDataset benchmark \cite{tudataset}, which contains different types of graphs, i.e., molecules and social networks, whose details can be shown in Table \ref{tab1}.
\paragraph{Implementation details.}
In our framework, we set the global view size to be 80$\%$ of the whole and the local view size to be 20$\%$ of the whole for molecular graphs, and 90$\%$ of the global view size and 10$\%$ of the local view size for social networks.
The measurement function between local views is composed of a 5-layer MLPs with batch normalization and RELU activation functions.
Its output is fed into a Sigmoid function, which outputs a scalar to indicate the similarity between two local views. 
\subsection{Unsupervised Representation Learning}
\paragraph{Experiments setting.}
We follow the work \cite{infograph} to evaluate the performance of the proposed method on unsupervised graph representation learning, where the model learns graph-level representations only through the supervision signals provided by the data itself without relying on labels.
After that, a SVM classifier is used to evaluate the quality of the representations.
In addition to the SOTA graph kernel based methods, WL \cite{wl}, DGK \cite{dgk}, for example, we also compare the proposed method with other eight advanced graph self-supervised learning methods, including node2vec \cite{node2vec}, sub2vec \cite{sub2vec}, graph2vec \cite{graph2vec}, Infograph \cite{infograph}, GraphCL \cite{graphcl}, JOAO \cite{joao} and SimGRACE \cite{simgrace}.
For our model, we adopt GIN as the encoder, and a sum pooling is used as the readout function.
We use 10-fold cross validation accuracy to report classification performance.
Experiments are repeated 5 times.
\begin{table*}[t]
    \centering
    \resizebox{\textwidth}{!}{
    \begin{tabular}{c|cccccccc}
        \toprule
        LR. & Method & NCI1 & PROTEINS & DD & COLLAB & RDT-B & RDT-M5K & AVG.\\
                 \midrule
                 \multirow{10}{*}{1$\%$} & No-pretrain & 60.72 $\pm$ 0.45 & - & - &  57.46 $\pm$ 0.25 & - & - & 59.09\\
                 & Augmentations & 60.49 $\pm$ 0.46 & - & - &  58.40 $\pm$ 0.97 & - & - &59.45 \\ 
                \cmidrule{2-9}
                 & GAE & 61.63 $\pm$ 0.84 & - & - &  63.20 $\pm$ 0.67 & - & - & 62.42 \\
                 & Infomax & 62.72 $\pm$ 0.65 & - & - &  61.70 $\pm$ 0.77 & - & - & 62.21 \\
                 & ContextPred & 61.21 $\pm$ 0.77 & - & - &  57.60 $\pm$ 2.07 & - & - & 59.41 \\
                 & GraphCL & 62.55 $\pm$ 0.86 & - & - &  64.57 $\pm$ 1.15 & - & - & 63.56 \\
                 & JOAO & 61.97 $\pm$ 0.72 & - & - &  63.71 $\pm$ 0.84 & - & - & 62.84 \\
                 & JOAO V2 & 62.52 $\pm$ 1.16 & - & - &  64.51 $\pm$ 2.21 & - & - & 63.52 \\
                 & SimGRACE & 64.21 $\pm$ 0.65 & - & - &  64.28 $\pm$ 0.98 & - & - & 64.25 \\
                 \cmidrule{2-9}
                 & MSSGCL & \textbf{64.63 $\pm$ 0.75} & - & - &  \textbf{65.02 $\pm$ 0.78} & - & - & \textbf{64.88} \\
                \midrule
                 \multirow{10}{*}{10$\%$} & No-pretrain & 73.72 $\pm$ 0.24 & 70.40 $\pm$ 1.54 & 73.56 $\pm$ 0.41 &  73.71 $\pm$ 0.27 & 86.83 $\pm$ 0.27 & 51.33 $\pm$ 0.44 & 71.56 \\
                        & Augmentations & 73.59 $\pm$ 0.32 & 70.29 $\pm$ 0.64 & 74.30 $\pm$ 0.81 &  74.19 $\pm$ 0.13 & 87.74 $\pm$ 0.39 & 52.01 $\pm$ 0.20 & 72.02\\ 
                \cmidrule{2-9}
                         & GAE & 74.36 $\pm$ 0.24 & 70.51 $\pm$ 0.17 & 74.54 $\pm$ 0.68 &  75.09 $\pm$ 0.19 & 87.69 $\pm$ 0.40 & 33.58 $\pm$ 0.13 & 69.30\\
                 & Infomax & \textbf{74.86 $\pm$ 0.26} & 72.27 $\pm$ 0.40 & 75.78 $\pm$ 0.34 &  73.76 $\pm$ 0.29 & 88.66 $\pm$ 0.95 & 53.61 $\pm$ 0.31 & 73.16 \\
                 & ContextPred & 73.00 $\pm$ 0.30 & 70.23 $\pm$ 0.63 & 74.66 $\pm$ 0.51 &  73.69 $\pm$ 0.37 & 84.76 $\pm$ 0.52 & 51.23 $\pm$ 0.84  & 71.26\\
                 & GraphCL & 73.63 $\pm$ 0.25 & 74.17 $\pm$ 0.34 & 76.17 $\pm$ 1.37 &  74.23 $\pm$ 0.21 & 89.11 $\pm$ 0.19 & 52.55 $\pm$ 0.45 & 73.48 \\
                 & JOAO & 74.48 $\pm$ 0.27 & 72.13 $\pm$ 0.92 & 75.69 $\pm$ 0.67 &  75.30 $\pm$ 0.32 & 88.14 $\pm$ 0.25 & 52.83 $\pm$ 0.54 & 73.10 \\
                 & JOAO V2 & 74.86 $\pm$ 0.39 & 73.31 $\pm$ 0.48 & 75.81 $\pm$ 0.73 &  75.53 $\pm$ 0.18 & 88.79 $\pm$ 0.65 & 52.71 $\pm$ 0.28 &73.50 \\
                 & SimGRACE & 74.60 $\pm$ 0.41 & 74.03 $\pm$ 0.51 & 76.48 $\pm$ 0.52 &  74.74 $\pm$ 0.28 & 88.96 $\pm$ 0.62 & 53.94 $\pm$ 0.64  & 73.78\\
                 \cmidrule{2-9}
                 & MSSGCL & 74.77 $\pm$ 0.31 & \textbf{75.76 $\pm$ 0.52} & \textbf{78.89 $\pm$ 0.18} &  \textbf{76.02 $\pm$ 0.13} & \textbf{90.58 $\pm$ 0.34} & \textbf{54.36 $\pm$ 0.24}  & \textbf{75.06} \\
        \bottomrule
    \end{tabular}
    }
    \caption{Comparison of classification accuracy with other baselines in semi-supervised 
    setting. AVG. denotes the average accuracy.}
    \label{tab3}
\end{table*}
\begin{table}[htbp]
    \centering
    \resizebox{\linewidth}{!}{
    \begin{tabular}{c|ccccc}
        \toprule
        Method & NCI1 & DD & COLLAB& RDT-B \\
        \midrule
        MSSGCL & 81.45$\pm$0.48 & 79.73$\pm$0.44 &73.40$\pm$0.72 & 91.08$\pm$0.78 \\ 
        w/o global-global & 80.27$\pm$0.51 & 78.62$\pm$0.49 &71.92$\pm$1.10 & 88.87$\pm$2.42 \\ 
        w/o global-local & 80.70$\pm$0.40 & 79.37$\pm$1.18 &73.08$\pm$0.71 & 89.75$\pm$1.00 \\ 
        w/o local-local & 80.74$\pm$0.51 & 79.45$\pm$0.40 &72.86$\pm$0.54 & 89.83$\pm$1.41 \\ 
        \bottomrule
    \end{tabular}
    }
    \caption{Ablation study on four benchmark datasets.}
    \label{tab4}
\end{table}
\paragraph{Results analysis.}
The results of the downstream graph classification task are shown in Table \ref{tab2}.
Although graph kernel-based methods can perform well on a single dataset, they cannot be extended to all datasets.
Similar to our method, GraphCL constructs comparison paths between small-scale subgraph pairs, but it ignores rich global information and cannot achieve better performance.
However, MSSGCL can achieve good performance on all datasets, outperforming all other baseline models.
This can be attributed to the fact that our method considers multi-scale views of the graph and combines multi-scale features according to the semantic relationships between views to form a favorable feature space.

\subsection{Semi-supervised Representation Learning}
\paragraph{Experiments setting.}
For semi-supervised setting, we pre-train a GNN in an unsupervised manner with all data, and then fine-tune the GNN with a certain percentage of labels on the same datasets.
Since the pre-training and fine-tuning of graph-level tasks in semi-supervised learning are less studied in the past, we additionally introduce several network embedding methods: GAE \cite{vgae}, local $\&$ global representation consistency enforcement \cite{dgi} and ContextPred \cite{contextpred}.
The rest of the baselines also include SOTA graph self-supervised learning methods, such as GraphCL \cite{graphcl}, JOAO \cite{joao} and SimGRACE \cite{simgrace}.
Following the settings in GraphCL, we employ 5-layer Residual Graph Convolutional Network (ResGCN) \cite{resgcn} with 128 hidden dimensions as our backbone network, and adopt 10-fold cross validation.
Experiments are repeated 5 times.
\paragraph{Results analysis.}
For the semi-supervised graph classification task, results are shown in Tabel \ref{tab3}, where two subtasks are reported with label rates of 1$\%$ and 10$\%$, respectively.
For 1$\%$ label rate setting, MSSGCL outperforms all the baseline models.
For 10$\%$ label rate setting, MSSGCL greatly outperforms the previous baselines and achieves the optimal performance on 6 out of 7 datasets.
Compared with GraphCL, which only considers single size subgraph, MSSGCL achieves an average 2$\%$ improvement.
\subsection{Ablation Study}
In this section, we will study three contrastive relations, i.e., global-global, global-local, and local-local.
We construct different variants by removing different loss terms to verify their effectiveness.

As can be seen from results in Table \ref{tab4}, when the model removes the global-global term, the performance drops significantly because the term contains rich global information.
When the local-local contrastive relationship is added to the loss term, the model performance can be improved to a certain extent, which clearly shows that our regressor is effective. 
And when the model considers three contrast relationships at the same time, the model performance can reach the optimum.
\begin{figure}[h!]
\vspace{-1em}
\centering
\subfigure[$\lambda _1$ and $\lambda _2$]{
\begin{minipage}[t]{0.48\linewidth}
\centering
\includegraphics[scale=0.24]{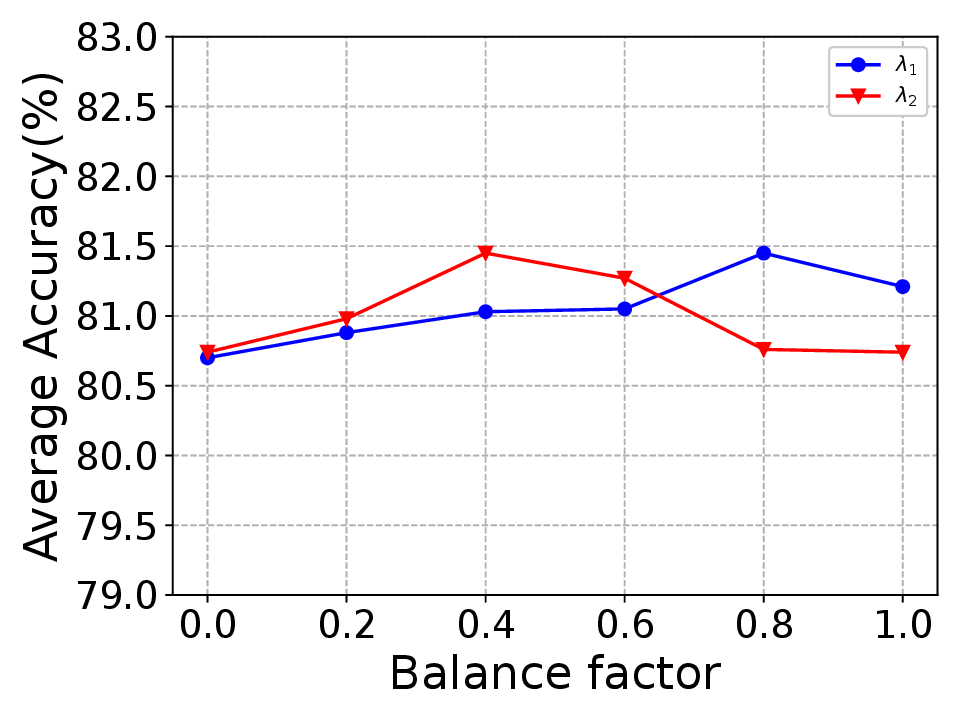}
\label{fig3a}
\end{minipage}%
}%
\subfigure[Batch Size]{
\begin{minipage}[t]{0.48\linewidth}
\centering
\includegraphics[scale=0.24]{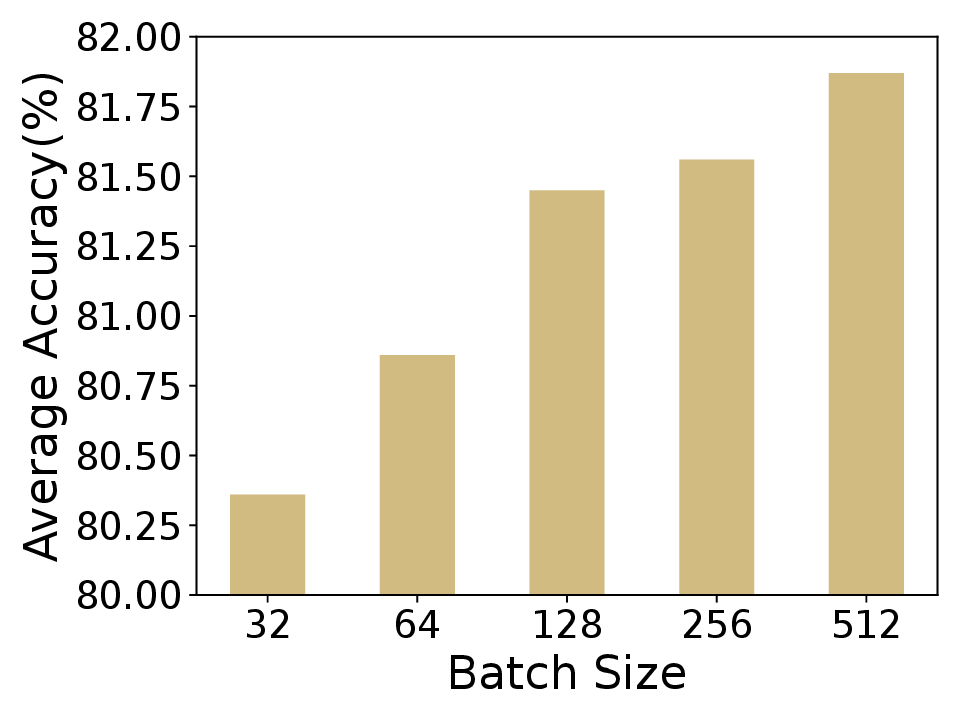}
\label{fig3b}
\end{minipage}%
}%

\centering
\subfigure[Epochs]{
\begin{minipage}[t]{0.48\linewidth}
\centering
\includegraphics[scale=0.24]{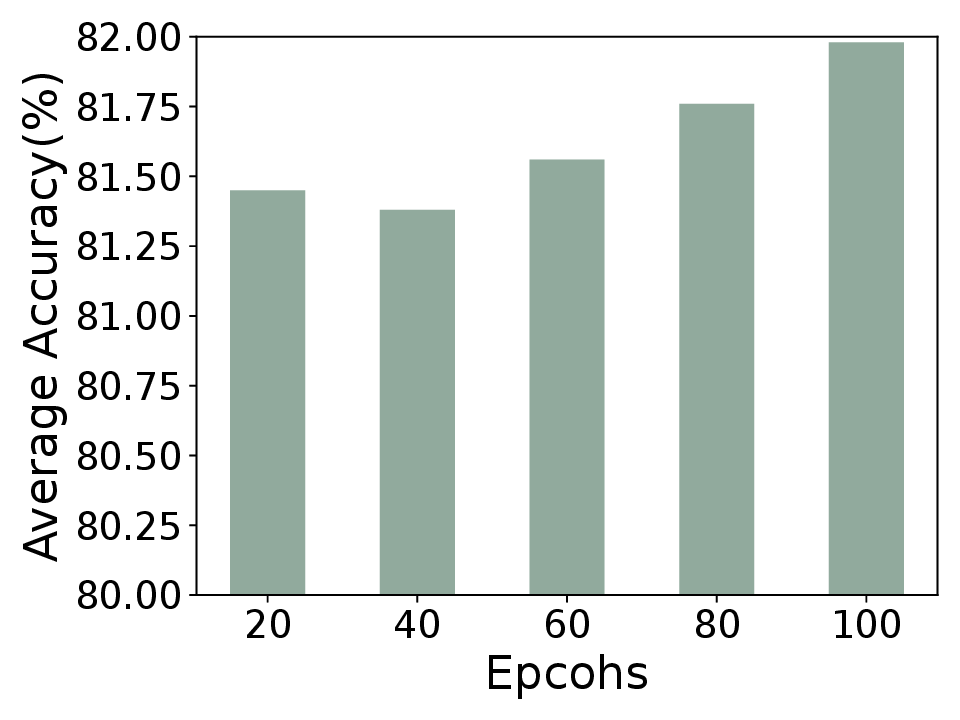}
\label{fig3c}
\end{minipage}%
}%
\subfigure[Hidden Size]{
\begin{minipage}[t]{0.49\linewidth}
\centering
\includegraphics[scale=0.24]{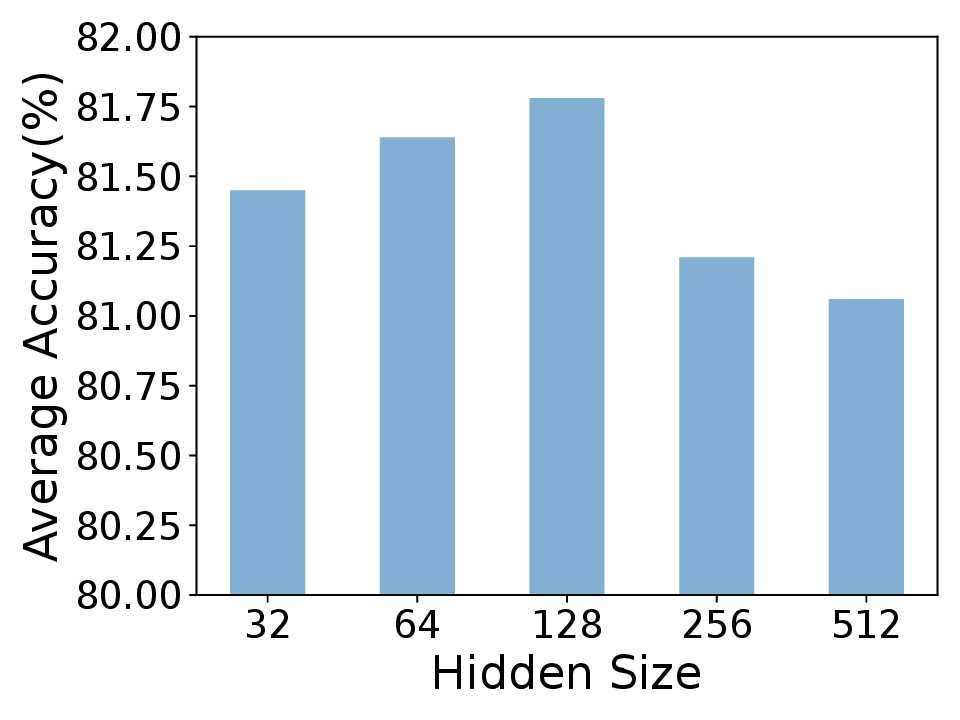}
\label{fig3d}
\end{minipage}%
}%
\caption{Classification accuracies of MSSGCL on NCI1 under different hyper-parameters.}
  \label{fig3}
\end{figure}
\subsection{Hyper-parameters Study}
\paragraph{Balance factor for the loss.}
In this section, we investigate the sensitivity of hyper-parameters $\lambda _1$ and $\lambda _2$.
Figure \ref{fig3a} shows the classification accuracy of MSSGCL with different values.
With the increase of $\lambda _1$, the performance of the model has been continuously improved, but it tends to be saturated after a certain level, which shows that establishing the connection between local and global can help the model learn better representations. 
On the other hand, properly encouraging the dissimilarity of local representations can promote the performance of the model, but when $\lambda _2$ is too large, the performance of the model will decrease. 
We believe that this is because there is still a low semantic similarity between local view pairs, excessive dissimilarity can impair the quality of learned representations.

\paragraph{Batch size, hidden size and epochs.}
Figure \ref{fig3b} and Figure \ref{fig3c} show the performance of MSSGCL at different batch sizes and epochs.
From results of two figures, it can be seen that larger batch size and training epochs lead to better performance, which is consistent with the findings of the work \cite{simclr}.
The possible reason is that larger batch size provides more samples for comparison.
Similarly, training longer time will generate more negative samples.
Figure \ref{fig3d} shows the sensitivity of the hidden size.
We can see that with the increasing of hidden dimension from 32 to 128, the performance gradually improves.
As the hidden dimension continues to increase, the performance begins to degrade, which may be caused by the model overfitting.
\section{Related Work}
\subsection{Graph Neural Network}
In recent years, graph neural networks have emerged as a promising method for analyzing graph due to their powerful expressive power.
They mainly follow the mechanism of message passing (or neighborhood aggregation) \cite{mpnn}.
Each node captures the attribute and structural information of neighbor nodes through message passing to update its own node representations, and then a shared linear transform is used to map the representations of nodes into a new feature space.
After the iteration of $k$-layer, the representation vectors of nodes can capture the information of $k$-hop neighbors.
Graph Convolutional Network (GCN) \cite{GCN} adopts the information of 1-hop neighbors to update node features, where the weight of each neighbors depends on the node degree.
Graph Attention Network (GAT) \cite{gat} considers the weight difference of neighbors through attention mechanism.
Graph Isomorphic Networks (GIN) \cite{gin}, inspired by the Weisfeiler-Lehman (WL) kernel, use simple summation operations and multilayer perceptrons (MLPs) to achieve the most powerful discriminative capabilities.
These methods mainly focus on supervised learning, which means that a large amount of labeled graph is required.
However, obtaining manually annotated labels is expensive in terms of time and labor, so our method mainly focuses on unsupervised/self-supervised learning.
\subsection{Graph Contrastive Learning}
Contrastive learning has been widely used in the field of computer vision with promising results.
Affected by this, some recent studies have begun to introduce contrastive learning into the field of graph learning.
The basic idea is to promote the embedding of augmentation views generated from the same instance to be closer, while those from different instances are opposite.
Encoders trained in this way can be used for downstream tasks.
DGI \cite{dgi} opens a precedent for graph contrastive learning, which treats node and graph-level representations as postive pairs and maximize their mutual information.
MVGRL \cite{mvgrl} further improves model performance by extending DGI to multiple views and cross-contrasting between them through graph diffusion.
Sub-Con \cite{subcon} learns node representations by sampling the subgraph and taking the central node and subgraph as a postive sample pair.
GraphCL\cite{graphcl} proposes four data augmentation methods for graphs, and proves subgraph sampling is an augmentation method beneficial to different types of datasets.
Cuco \cite{cuco} combines curriculum learning with contrastive learning, and proposes a scoring and a pacing functions to automatically select negative samples during training.

Although the above methods have made good progress in GCL, they ignore the semantic association between augmented views, while our method can adopt different learning strategies according to the semantic information of views at different scales.

\section{Conclusion}
In this paper, we investigate the semantic association among subgraphs at different scales, and propose a novel multi-scale subgraph contrastive learning method.
Based on the semantic association, we define two different types of subgraph, i.e., global view and local view.
We construct a variety of contrastive relations between views, and implement different learning strategies to achieve mutual matching of semantic information between augmented views.
We conduct graph classification experiments on eight real-world datasets, and the experimental results demonstrate that the proposed method can outperform the state-of-the-arts in unsupervised and semi-supervised learning.

\section*{Acknowledgments}
This work is supported in part by the National Natural Science Foundation of China (No.62172052, 61901297), the Special Foundation for Beijing Tianjin Hebei Basic Research Cooperation (J210008, 21JCZXJC00170, H2021202008), and The Open Project of Anhui Provincial Key Laboratory of Multimodal Cognitive Computation, Anhui University (No. MMC20210).


\bibliographystyle{named}
\bibliography{ijcai23}

\end{document}